\documentclass[journal]{IEEEtran}
\usepackage{authblk}
\usepackage{graphicx} % DO NOT CHANGE THIS
\usepackage{multicol,amsmath,amsfonts,amssymb,amsthm}
\usepackage[linesnumbered,boxed]{algorithm2e}
\DontPrintSemicolon
\usepackage{enumerate,etoolbox,paralist,url,makecell,xcolor}
\usepackage{booktabs}       % professional-quality tables
\SetKwInOut{Header}{Algorithm1:}
\SetKwInOut{Input}{input}
\SetKwInOut{Output}{output}
\SetFuncSty{texttt}
\SetKwFunction{FCtr}{constructImage}
\SetKwProg{Fn}{function}{:}{}
\begin{document}
\title{Toward an ImageNet Library of Functions\\ for Global Optimization Benchmarking}
\author{Boris Yazmir} 
    \affil{Migal Institute, Upper Galilee, Israel}
\author{Ofer M.~Shir} 
    \affil{Computer Science Department, Tel-Hai College, and Migal Institute, Upper Galilee, Israel}
\maketitle
     % Affiliations
%     \textsuperscript{\rm 1} Migal Institute, Upper Galilee, Israel\\
%     \textsuperscript{\rm 2} Computer Science Department, Tel-Hai College, Israel\\
%     borisy@migal.org.il, ofersh@telhai.ac.il

\newcommand{\an}[1]{\textcolor{blue}{#1}}
\newcommand{\td}[1]{\textcolor{red}{#1}}

\begin{abstract}
Knowledge of search-landscape features of Black-Box Optimization (BBO) problems offers valuable information in light of the Algorithm Selection and/or Configuration problems.
Exploratory Landscape Analysis (ELA) models have gained success in identifying \textit{predefined human-derived} features and in facilitating portfolio selectors to address those challenges. 
Unlike ELA approaches, the current study proposes to transform the identification problem into an image recognition problem, with a potential to detect \textit{conception-free, machine-driven} landscape features.
To this end, we introduce the notion of Landscape Images, which enables us to generate imagery instances per a benchmark function, and then target the classification challenge over a diverse generalized dataset of functions.
%We propose an identification framework for $d$-dimensional continuous BBO test-suites, provided as input with a set of $N$ search-points, sampled at random, together with their query/function values.
We address it as a supervised multi-class image recognition problem and apply basic artificial neural network models to solve it.  %(used previously for digits recognition) 
The efficacy of our approach is numerically validated on the noise-free BBOB and IOHprofiler benchmarking suites.
%All test-functions, at high dimensional spaces ($\lesssim 40$), are demonstrated to be correctly classified at high precision rates ($\approx$90\%) when $N$ is in the order of $d$.
This evident successful learning is another step toward automated feature extraction and local structure deduction of BBO problems.
By using this definition of landscape images, and by capitalizing on existing capabilities of image recognition algorithms, we foresee the construction of an ImageNet-like library of functions for training generalized detectors that rely on machine-driven features.
\end{abstract}

%\keywords{Black-Box Optimization, Algorithm Selection, Algorithm Configuration, noiseless COCO/BBOB, LeNet5 convolutional neural network}

\section{Introduction}\label{sec:introduction}
Global Optimization of continuous domains \cite{globaloptimization} constitutes an underlying core problem of many disciplines, including Artificial Intelligence and Machine Learning, and at the same time lies in the heart of many application areas. 
Its focal point is an objective function, $f~:~\mathcal{S}\subseteq \mathbb{R}^d \rightarrow\mathbb{R}$, where $\mathcal{S}$ denotes the feasible region as defined by a given set of constraints. 
When the analytical forms of the objective function and the constraints are available, the problem is considered to be White-Box Optimization (WBO) and may be explicitly treated in a bottom-up fashion, by exploiting the explicit problem-structure and the available instance-data. Furthermore, when $f$ is convex and $\mathcal{S}$ satisfies certain conditions, the problem is guaranteed to be solvable by exact solvers \cite{Boyd}.
An optimization problem is defined as a Black-Box Optimization (BBO) problem when there is strictly zero information about $f$. 
In principle, this definition may be extended also to some forms of noise/uncertainty, at multiple levels, with unknown distributions. However, we exclude noisy problems from our current consideration, and therefore it is assumed that all BBO problems herein are noiseless.
Randomized search heuristics are popular techniques for treating BBO problems \cite{Wegener2004}. 
They operate in a trial-and-error fashion, by evaluating candidate solutions and using their function values to evolve the strategy upon which the next \emph{search-points} are drawn. 
They serve as fine alternatives to exact solvers, particularly on non-convex search landscapes. Especially, Evolution Strategies \cite{ESchapter2018,EStoRL} are the predominant heuristics that excel on continuous search landscapes, with the CMA-ES being the state-of-the-art \cite{Baeck2013contemporary}.

\subsection{The Algorithm Selection and Configuration Problems} 
The \textit{algorithm selection problem} \cite{RICE1976} is defined as the task to choose the `fittest' algorithm, within an available portfolio, per a given problem-instance.
At the same time, the \textit{algorithm configuration problem} refers to the task of choosing the `fittest' parameter settings for a family of optimization problems, as defined by a distribution of problem-instances. 
Notably, they both constitute meta-optimization problems \cite{Hoos2012}, where the empirical performance measures serve as the objective function values(s) (and also refer to the aforementioned `fittest' term).
The selection problem has been widely investigated | a large variety of approaches were defined (e.g., ranking-based selection \cite{AlgorithmSelection2015}, or scheduling \cite{LindauerBH16}), and the so-called ICON challenge \cite{ICON_2017} has been defined as an AI benchmarking.
The configuration problem has also been extensively explored. 
Prominent approaches are the Per-Instance Algorithm Configuration (PIAC) \cite{PIACS_Hoos2006}, initially defined for Combinatorial Optimization problems, and the Instance-Specific Algorithm Configuration (ISAC) \cite{ISAC2010}. They were both successfully transferred into the BBO continuous domain (see \cite{Schoenauer2017} utilizing PIAC with the CMA-ES, and \cite{Tinus2013} for the ISAC). 
The following step was to consider selection and/or configuration problems per \textit{generalized problem features} \cite{Tinus2013}, and indeed, feature-based algorithm configuration on continuous BBO was accomplished using the CMA-ES \cite{Schoenauer2017}.
However, the success of existing approaches is dependent upon informative instance features. In other words, the problem of the learning model to choose the `fittest' algorithm on uninformative instance features remains an open challenge, with ELA-based approaches exhibiting success (see Section \ref{sec:ELA}). This open challenge provides us with the motivation to obtain the \textbf{machine capability to extract BBO features and deduce local structure using a small number of queries and without relying on human conceptions for defining the features}.
In the BBO context, we are interested in the online task of choosing the `fittest' optimizer/solver, per each search-phase during the optimization run, by letting the machine recognize the local structure with a small number of queries.

\subsection{BBO Benchmarking}
Benchmarking aims at supporting practitioners in choosing the ``best'' algorithmic technique and its fittest instantiation for the optimization problem at hand through a systematic empirical investigation and comparison amongst competing techniques. Benchmarking is typically carried out in light of research questions such as \textit{how does the algorithm perform on different classes of optimization problems}, \textit{which \textbf{optimization problem features} possess the strongest impact on the accuracy and/or the convergence speed}, \textit{how does the performance scale with increasing problem complexity}, etc. \cite{ShirDB18}.
The continuous BBO Benchmarking suite (BBOB) for continuous search \cite{BBOB_GECCO2010}, which is realized by the so-called COCO platform, constitutes an established testing framework for evaluating the performance of continuous optimizers. The noise-free suite encompasses 24 basis functions classified as follows \cite{Hansen2014Real}: 
(i) Separable functions, (ii) Functions with low or moderate conditioning, (iii) Functions with high conditioning and unimodal, (iv) Multi-modal functions with adequate global structure, and (v) Multi-modal functions with weak global structure.
The COCO procedure generates random instances of these basis functions, based upon translations and rotations.
In the context of the configuration problem, \cite{Schoenauer2016} accomplished a per-instance configuration over the BBOB test functions using the CMA-ES algorithm.

%\td{TODO: IOHexperimenter intro}
The recently announced IOHprofiler platform \cite{DoerrYHWSB20} provides a competing environment for automated execution of benchmarking (by means of its IOHexperimenter component) and offers analyses and visualizations by means of its IOHanalyzer component. 
In this work we additionally consider the 23 discrete optimization problems that are proposed by the IOHexperimenter.

\subsection{Fitness Landscapes}\label{sec:fitness_landscapes}
The concept of a \textit{fitness landscape} constitutes an underlying foundation of several disciplines \cite{Stadler2002} -- Physics, Evolutionary Biology, and Global Optimization, to name the predominant -- with practical implications, e.g., in Control \cite{Rabitz2014}.
A fitness landscape is defined as a mapping from a configuration space (having a predefined notion of distance) into $\mathbb{R}$. A common goal is to characterize a given landscape, in order to arrive at its critical points. This goal is translated to unveiling certain mathematical properties of the landscape, entitled features. 
In the continuous case, given a BBO function $f: \mathbb{R}^d \to \mathbb{R}$, a feature is to be statistically/empirically extracted from samples of $f^2$ -- that is, a population of pairs $\left\{\vec{x}_i, f\left(\vec{x}_i \right)\right\}$. Explicitly, the set $\left\{f\left(\vec{x}_i \right)\right\}$ is to be denoted by $\mathcal{Y}$, and certain properties are to be extracted \cite{ELA_gecco2010,Schoenauer2016}:
$y$-Distribution features stem from the distribution of $\mathcal{Y}$'s values, Levelset features stem from their relative ranking, Curvature features correspond to numerically estimated gradients and Hessian matrices, and so on.
Automatically characterizing fitness landscapes, in the BBO context, has been a long lasting challenge (see \cite{FitnessLandscapeCharacter_2013} for a broad review and a survey of techniques), enjoying also some theoretical background (see, e.g., \cite{Jansen2001}).

\subsection{Exploratory Landscape Analyses}\label{sec:ELA}
An established approach of fitness landscapes' characterization is the so-called Exploratory Landscape Analysis (ELA), which is rooted in discrete optimization \cite{DiscreteLandscapeCharac_2009}. 
In the past decade, a prominent continuous optimization methodology emerged \cite{ELA_gecco2010}. 
Its pillar is a predefined set of 55 features, which has been defined by global optimization experts. 
By conducting systematic sampling and applying machine learning, those features may be recognized when given a BBO.
Furthermore, advanced ELA research yielded extensions (see, e.g., \cite{ELA-Continuous_2014}) and eventually led to an automated technique that subsequently addresses the Algorithm Selection Problem \cite{ELA+ASP_2019} in a direct manner.

\subsection{Other Work}
A recent, AI-oriented, study addressed an equivalent target in WBO \cite{LoreggiaMSS_aaai16}: It introduced a learning procedure capable of correctly classifying a given WBO problem-instance and assigning an algorithm portfolio to solve it.
The assumed input for that procedure was an ASCII-based textual description of the Mathematical Programming model (that is, a conventional WBO encoding of the explicit objective function and constraints by means of a text file), where the solution's core idea was to transform this textual information into an image, and classify it using Deep Learning. 
Except for the obvious difference that stem from the distinction between WBO and BBO, i.e., the availability of an optimization model versus the necessity to sample the black-box by queries, our study differs also in its underlying learning target -- we aim to learn the search landscape features, and to this end we will define the Landscape Images in Section \ref{sec:method}.

\subsection{Core Idea, Aims and Contribution}
% This part should describe idea behind use of landscape images
The \textit{long-range aim} of this study is to develop machine-driven capability to recognize the local structure of search landscapes. 
To this end, our idea is to recognize test-functions' type based on the embedding of their input and output.
Computer vision and specifically image recognition are well established fields that flourished with the rise and the advancement of deep learning methods \cite{8320684}. 
The novel perspective of this study is to view the local structure recognition problem as a topology pattern recognition problem, and accordingly, let the machine derive the characteristic features.
Therefore, we propose to convert the functions' detection problem into an image recognition problem and address it using standard image recognition techniques. 
The core of the method is the structural organization/embedding of the BBO input-output couples within synthetic images, to which we refer herein as \textbf{Landscape Images}. 

The \textit{specific target} of this work is to evaluate the possibility of local structure identification using image recognition techniques, and \textbf{without using human-derived features}. 
We also explore the impact of different structural configurations in landscape images.
The concrete contributions of this paper are the following:
\begin{itemize}
    \item Encapsulating high-dimensional continuous BBO search-points within a so-called image landscape is an effective, compact form of representing local information of the landscape.
    \item Accurately classifying such Image Landscapes per their generating functions is translated into an image recognition problem.
    \item Solving this problem is accomplished using basic artificial neural network models, and numerically validated on the noise-free BBOB and IOHexperimenter benchmarking suites.
    \item In practice, given an unknown objective function (BBO), the established model will be able to recognize the closest image, within the set of known objective functions' images, with the strongest resemblance.  
    Such a set of known images will play the equivalent role of ImageNet for optimization problems' detection. 
    Subsequently, it will be possible to fit an algorithm to a given problem at hand, which in turn will enable smoother problem-solving.
\end{itemize}

%I think it worth to add here sentence that states clearly what are the topics of this specific article in our frame work.BY We should say that we want to recognize functions,

%\subsection{Paper structure}
% It should be part of introduction section.
The remainder of this paper is organized as follows.
%The problem is formally stated in Section \ref{sec:problem}, where the practical work-flow is described in detail.
Next, Section \ref{sec:method} describes the notation and defines the main components that comprise the proposed approach. It also formally states the targeted problem and then summarizes the taken approach. 
In Section \ref{sec:results} we specify our experimental setup and present the numerical observations.
Finally, Section \ref{sec:discussions} discusses the results, provides concluding remarks and outlines directions for future work.

% We should say something about function detection importance for engineering.

%\section{Automated Black-Box Function Detection}\label{sec:problem}
\section{Landscape Images Definition, Problem Formulation \& Approach}\label{sec:method}

\subsection{Notation}
Given an optimization benchmarking test-suite, let 
$${f_k}: \mathbb{R}^d \to \mathbb{R},~k=1,2,\ldots,F$$ 
denote the $d$-dimensional $k^{th}$ objective function, subject to minimization. 
Importantly, each function is instantiated using randomized mathematical operations (e.g., rotations and translations); \textbf{we denote by $\varphi_{k,i}$ a random instance of $f_k$}.\\
A \textit{sample vector} is defined as some form of concatenation of a random search-point, $\vec{x}^{(j)}$, with its associated objective function value, $\varphi_{k,i}\left(\vec{x}^{(j)}\right)$, denoted altogether as $\vec{\tau}^{(k,i,j)}$. 
An example for such a concatenation, which may be realized in multiple fashions, is a $d^{\prime}$-dimensional vector ($d^{\prime}\equiv d+1$) of the form 
$$\vec{\tau}^{(k,i,j)}(\vec{x}^{(j)}):= \left[\vec{x}^{(j)~T},\varphi_{k,i}\left(\vec{x}^{(j)}\right) \right]\in \mathbb{R}^{d^{\prime}}. $$
% j - number of xFx vector
%Let $N$ denote the  of a set of sample vectors. 
\subsection{Landscape Images}
A landscape image is a data structure of sample vectors. 
We first define a \textit{core image} $\Psi^{(k,i)}$ %of the \textit{landscape image} 
as the aggregation of $N$ sample vectors, which are drawn with respect to $\varphi_{k,i}$. $N$ is the number of randomly sampled $\vec{x}^{(j)}$ vectors, drawn from a uniform distribution $\mathcal{U}(0, 1)$, and is entitled the fixed \textit{sample-size}:
$$\Psi^{(k,i)}:= \left[ \vec{\tau}^{(k,i,1)}(\vec{x}^{(1)}),\ldots,\vec{\tau}^{(k,i,N)}(\vec{x}^{(N)}) \right]\in \mathbb{R}^{N \times d^{\prime}}. $$
Then, a \textit{landscape image} $\Gamma^{(k,i)}$ is defined as some \textit{square embedding} of a core image within a fixed, framework size $M$, to yield an $M\times M$ instance. 
The concrete embedding subscribes to one of the types and forms that are listed below. Following this listing, we will address the issue of sizing $M$.
%This core image is embedded into the main landscape image structure. 
%\paragraph{Sizing the Image}
% Main images are of size of 32x32 $(M:=32)$, as used frequently in computer vision problems and was used in LeNet5 neural network \cite{Lecun98gradient-basedlearning}. Here we will describe different variations of the main landscape image structures.

\begin{figure*}
\centering
\includegraphics[width=1\textwidth]{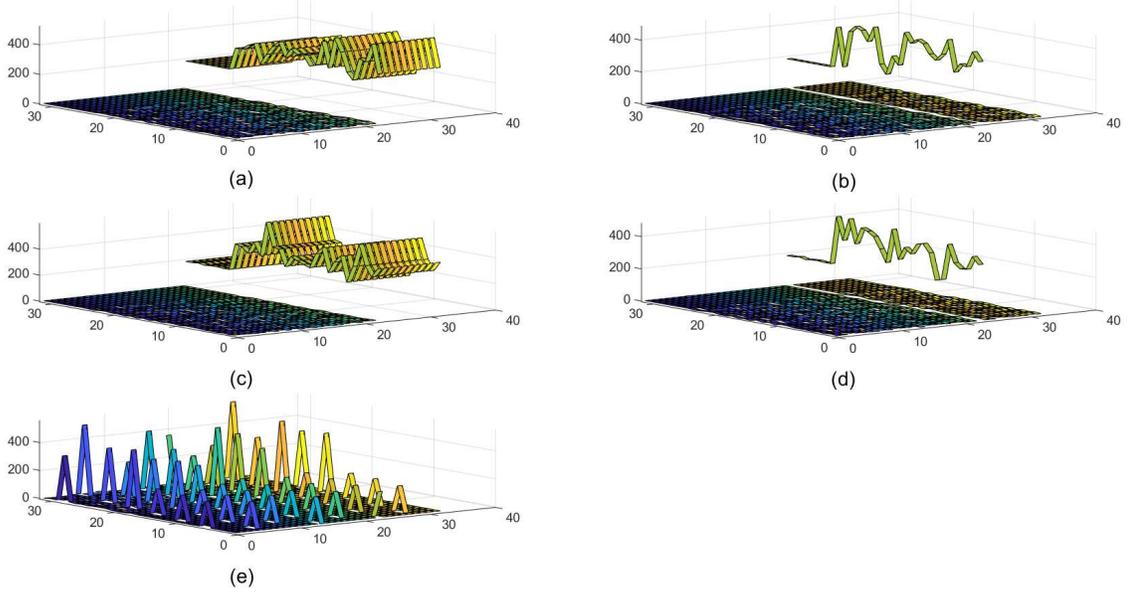} 
 \caption{Example of \textbf{a} - type 1, \textbf{b} - type 2' \textbf{c} - type 3, \textbf{d} - type 4 and \textbf{e} - type 5 landscape images for BBOB Sphere function at a random location with $d=22$ and $N=24$.\label{fig:Landim1}}
\end{figure*}

\paragraph{Landscape Image Type-1.} A landscape image with a replication of $\varphi_{k,i}\left(\vec{x}^{(j)}\right)$ and the use of $d$-dimensional zero vectors. This is the basic structure of a landscape image. Each image row is composed of $\vec{\tau}^{(k,i,j)}$ and $M-d-1$ replications of $\varphi_{k,i}\left(\vec{x}^{(j)}\right)$. 
$M-N$ zero vectors, $\left\{\vec{x}^{(j)}:=\vec{0}\right\}_{j=N+1}^{M}$, and their corresponding function values $\varphi_{k,i}\left(\vec{0}\right)$, are added below the first $N$ rows (that represent the randomly sampled vectors $\left\{\vec{x}^{(j)}\right\}_{j=1}^{N}$). Those zero vectors are added to cope with random translations of instances of the objective function $\varphi_{k,i}$. Figure \ref{fig:Landim1} depicts an example of type 1-5 landscape images for the BBOB Sphere function at a random location with $d=22$ and $N=24$.
A schematic description of this image type  $\Gamma_1^{(k,i)}$ is prescribed below: 
%This image structure can get $\vec{\tau}^{(k,i,j)}$ with dimension ups to $M=d+1$. 
$$
\begin{Bmatrix}
\vec{\tau}^{(k,i,1)}\left(\vec{x}^{(1)}\right)_{1\times d^{\prime}}& \varphi_{k,i}\left(\vec{x}^{(1)}\right)_{(M-d-1)\textrm{ replications}}\\
\ldots & \ldots\\
\ldots & \ldots\\
\vec{\tau}^{(k,i,N)}\left(\vec{x}^{(N)}\right)_{1\times d^{\prime}} & \varphi_{k,i}\left(\vec{x}^{(N)}\right)_{(M-d-1)\textrm{ replications}}\\
\vec{\tau}^{(k,i,N+1)}\left(\vec{0}\right)_{1\times d^{\prime}} & \varphi_{k,i}\left(\vec{0}\right)_{(M-d-1)\textrm{ replications}}\\
\ldots & \ldots\\
\ldots & \ldots\\
\vec{\tau}^{(k,i,M)}\left(\vec{0}\right)_{1\times d^{\prime}} & \varphi_{k,i}\left(\vec{0}\right)_{(M-d-1)\textrm{ replications}}\\
\end{Bmatrix}
$$

% \begin{figure}
% \centering
% \includegraphics[width=1\columnwidth]{LandIm.eps} 
%  \caption{Example of type 1 landscape image for BBOB Sphere function at a random location with $d=22$ and $N=24$.\label{fig:Landim1}}
% \end{figure}

\paragraph{Landscape Image Type-2.}
A landscape image with a replication of $\varphi_{k,i}\left(\vec{x}^{(j)}\right)$ and $\vec{x}^{(j)}$ together. $d$-dimensional zero vectors are used similarly to Type-1. Sample vectors $\vec{\tau}^{(k,i,j)}$ are replicated as much as possible at the same row. If the last replication exceeds the image size, as many vector components as possible are written in practice. A schematic description of this image type  $\Gamma_2^{(k,i)}$ is prescribed below:
$$
\begin{Bmatrix}
\vec{\tau}^{(k,i,1)}\left(\vec{x}^{(1)}\right)_{(M-d-1)/d^{\prime}\textrm{ replications}}\\
\ldots\\
\ldots\\
\vec{\tau}^{(k,i,N)}\left(\vec{x}^{(N)}\right)_{(M-d-1)/d^{\prime}\textrm{ replications}}\\
\vec{\tau}^{(k,i,N+1)}\left(\vec{0}\right)_{(M-d-1)/d^{\prime}\textrm{ replications}}\\
\ldots\\
\ldots\\
\vec{\tau}^{(k,i,M)}\left(\vec{0}\right)_{(M-d-1)/d^{\prime}\textrm{ replications}}\\
\end{Bmatrix}
$$
\paragraph{Landscape Image Type-3.} A landscape image with a replication of $\varphi_{k,i}\left(\vec{x}^{(j)}\right)$ and the use of $d$-dimensional zero and unit base vectors. 
This landscape image is similar to Type-1, with the addition of the unit base vectors, e.g., $\vec{e}_1:=[1,0,\ldots,0]^T,~\vec{e}_2:=[0,1,0,\ldots,0]^T,\ldots,\vec{e}_d:=[0,\ldots,0,1]^T$, which replace some of the replicated zero vectors. 
Those unit vectors are added to cope with random rotations of instances of the objective function $\varphi_{k,i}$. 
$M-N$ zero and unit vectors, $\left\{\vec{x}^{(j)} \in \left\{\vec{0},~\vec{e}_1,~\ldots,\vec{e}_d\right\} \right\}_{j=N+1}^{M}$ and their objective function values are added below the first $N$ rows. 
In case $d^{\prime}>M-N$, only the first $M-N$ zero and unit vectors are added. Otherwise, zero and unit vectors are replicated as much as possible. 
A schematic description of this image type  $\Gamma_3^{(k,i)}$ is prescribed below: 
$$
\begin{Bmatrix}
\vec{\tau}^{(k,i,1)}\left(\vec{x}^{(1)}\right)_{1\times d^{\prime}} & \varphi_{k,i}\left(\vec{x}^{(1)}\right)_{(M-d-1)\textrm{ replications}}\\
\ldots & \ldots\\
\ldots & \ldots\\
\vec{\tau}^{(k,i,N)}\left(\vec{x}^{(N)}\right)_{1\times d^{\prime}} & \varphi_{k,i}\left(\vec{x}^{(N)}\right)_{(M-d-1)\textrm{ replications}}\\
\vec{\tau}^{(k,i,N+1)}\left(\vec{0}\right)_{1\times d^{\prime}} & \varphi_{k,i}\left(\vec{0}\right)_{(M-d-1)\textrm{ replications}}\\
\vec{\tau}^{(k,i,N+2)}\left(\vec{e}_{*}\right)_{1\times d^{\prime}} & \varphi_{k,i}\left(\vec{e}_{*}\right)_{(M-d-1)\textrm{ replications}}\\
\ldots & \ldots\\
\ldots & \ldots\\
\vec{\tau}^{(k,i,*)}\left(\vec{0}\right)_{1\times d^{\prime}} & \varphi_{k,i}\left(\vec{0}\right)_{(M-d-1)\textrm{ replications}}\\
\vec{\tau}^{(k,i,*)}\left(\vec{e}_{*}\right)_{1\times d^{\prime}} & \varphi_{k,i}\left(\vec{e}_{*}\right)_{(M-d-1)\textrm{ replications}}\\
\ldots & \ldots\\
\end{Bmatrix}
$$
\paragraph{Landscape Image Type-4.}
A landscape image with a replication of $\varphi_{k,i}\left(\vec{x}^{(j)}\right)$ and $\vec{x}^{(j)}$ together. $d$-dimensional zero and unit vectors are used similarly to Type-3. Sample vectors $\vec{\tau}^{(k,i,j)}$ are replicated as much as possible at the same row, like in Type-2. If the last replication exceeds the image size, as many vector components as possible are written in practice. 
The prescription of  $\Gamma_4^{(k,i)}$ is given below: 
$$
\begin{Bmatrix}
\vec{\tau}^{(k,i,1)}\left(\vec{x}^{(1)}\right)_{(M-d-1)/d^{\prime}\textrm{ replications}}\\
\ldots\\
\ldots\\
\vec{\tau}^{(k,i,N)}\left(\vec{x}^{(N)}\right)_{(M-d-1)/d^{\prime}\textrm{ replications}}\\
\vec{\tau}^{(k,i,N+1)}\left(\vec{0}\right)_{(M-d-1)/d^{\prime}\textrm{ replications}}\\
\vec{\tau}^{(k,i,N+2)}\left(\vec{e}_{*}\right)_{(M-d-1)/d^{\prime}\textrm{ replications}}\\
\ldots\\
\ldots\\
\vec{\tau}^{(k,i,*)}\left(\vec{0}\right)_{(M-d-1)/d^{\prime}\textrm{ replications}}\\
\vec{\tau}^{(k,i,*)}\left(\vec{e}_{*}\right)_{(M-d-1)/d^{\prime}\textrm{ replications}}\\
\ldots\\
\end{Bmatrix}
$$
\paragraph{Landscape Image Type-5.} %Any vector length
A landscape image with different concatenated sample vectors $\vec{\tau}^{(k,i,j)}$. Initially concatenated sample vectors $\vec{\tau}^{(k,i,j)}$ are arranged in a vector of ${M}\times{M}$ length, denoted as $\vec{\eta}_{[1 \times (M \cdot M)]}$. 
It begins with a zero and a unit vector, and followed by randomly generated sample vectors. 
As many different sample vector $\vec{\tau}^{(k,i,j)}$ as possible are then added:
$$\vec{\eta} = \begin{bmatrix}
  \vec{\tau}^{(k,i,1)}\left(\vec{0}\right) & \vec{\tau}^{(k,i,2)}\left(\vec{e}_{*}\right) & \ldots & \vec{\tau}^{(k,i,\ell)}\left(\vec{x}^{(j)}\right) & \ldots
\end{bmatrix}$$
If the last $\vec{\tau}^{(k,i,j)}$ exceeds the size of $\vec{\eta}$, as many vector components of $\vec{\tau}^{(k,i,j)}$ as possible are written in practice.
Next, $\vec{\eta}$ is \textit{reshaped}, in a row-wise fashion, into an $M \times M$ image. 

\subsubsection*{Sizing the Images}
Importantly, in terms of capacity, given $M\times M$ landscape images of types 1-4, they are capable of capturing sample vectors of dimension $\bar{d^{\prime}}:=M$ %$(d^{\prime}=d+1)$ 
at most.
In contrast, Type-5 is designed to capture sample vectors of dimensions up to $\bar{d^{\prime}}_5 \lesssim {M}\times{M}$.\\
In the current study, the landscape images are of size $32\times 32$, as commonly utilized in the Computer Vision domain. 
The experimentation herein is thus limited to dimension $\bar{d}=31$ when using types 1-4. 
The extension to higher dimensions is easily done by setting $M$ appropriately and following the aforementioned construction protocols.

\subsection{Problem Formulation and Taken Approach}\label{sec:algorithm}
We target the following problem:
%\begin{quote}
Learn a model that correctly classifies a given landscape image $\Gamma^{(k,i)}$ to its underlying function identity $k$. 
%\end{quote}
It is essentially a multi-class identification problem, where each class is randomly instantiated throughout the datasets. By utilizing the landscape images we converted a problem of $1 \times d$ input - scalar output function mapping recognition, into an image recognition problem. This problem is formulated using a standard pseudo-code in Algorithm \ref{algo:highlevel}. It relies on the auxiliary function \texttt{constructImage} (Algorithm \ref{algo:constructImage}), as well as on some polymorphic definition of the so-called \textit{type trait} (to represent the images' types; not included).
\textbf{Importantly, constructing each image landscape requires at most $N+2$ objective function calls/queries}.

In practice, to address this problem, we can apply well-established image recognition techniques designated to detect the object responsible for generating the image. This object can be thus defined as the generator function. 
For example, given an image of a fox, the fox object can be viewed as a generator function providing a specific image topology.

\textbf{Convolutional Neural Networks} are widely used nowadays and are considered to be the state-of-the-art in the image recognition field \cite{8320684}. 
The efficiency of the learning process depends not only on the level of the network's complexity, but also on the effectiveness of data structure. 
Consequently, the relatively simple yet powerful, LeNet5 convolutional neural network (CNN), which dates back to 1998 \cite{Lecun98gradient-basedlearning}, was employed. This CNN proved to be efficient in recognition of handwritten characters in $32 \times 32$ images. LeNet5 comprises 7 different layers including convolution, pooling and fully connected layers \cite{Lecun98gradient-basedlearning}. 

%Add network image
%We should say that we use only one instatance of function to simplify.
% Start simple

\IncMargin{1em}
\begin{algorithm}
\caption{High-level description of the proposed approach. \texttt{generateInstance} of line-4 represents a selected procedure to instantiate the $k^{th}$ function using randomized mathematical operations.
\texttt{constructImage} of line-5 constitutes the core generation of the Landscape Images, and is independently described in Algorithm \ref{algo:constructImage}.\label{algo:highlevel}}
\Input{a benchmarking suite generator $\mathcal{G}$, dimensionality $d$,\\sample-size $N$,\\type trait $t$,\\training set's size per function $L$\\
}
\Output{a trained model $\mathcal{M}$ encompassing landscape-adapted features of $\mathcal{G}$ }
$\mathcal{D},\mathcal{T} \leftarrow [~],[~]$; //lists for landscape images and tags\;
\For{$k\leftarrow 1$ \KwTo $\left|F\right|$}{
\For{$\ell\leftarrow 1$ \KwTo $L$}{ %//some integer representing the number of trained instances
$\varphi_{\ell}\leftarrow\mathcal{G}.\texttt{generateInstance}\left(k \right)$\;
$\Gamma^{(k,\ell)} \leftarrow \texttt{constructImage}\left(d,~N,~\varphi_{\ell},~t\right)$\;
$\mathcal{D} \leftarrow \mathcal{D} \texttt{.append} \left(\Gamma^{(k,\ell)} \right)$\;
$\mathcal{T} \leftarrow \mathcal{T} \texttt{.append} \left(k \right)$\;
}
}
$\mathcal{D}_{\textrm{train}},\mathcal{T}_{\textrm{train}}\leftarrow \texttt{shuffleSyncLists}\left(\mathcal{D},\mathcal{T} \right)$\;
$\mathcal{M}_0 \leftarrow \texttt{initializeNN}\left( \right)$\;
$\mathcal{M}\leftarrow \texttt{trainNN}\left[ \mathcal{M}_0,~\left(\mathcal{D}_{\textrm{train}},\mathcal{T}_{\textrm{train}}\right) \right]$\;
%$\texttt{crossValidate}\left[ \mathcal{M},~\left(\mathcal{D}_{\textrm{test}},\mathcal{T}_{\textrm{test}}\right) \right]$\;
\Return{ $\mathcal{M}$ }\;
\end{algorithm}
\DecMargin{1em}
\IncMargin{2em}
\begin{algorithm*}
%\Fn{\FCtr{\textrm{dim}~d,~\textrm{size}~N,~\textrm{instance}~\varphi,~\textrm{type\_trait}~t}}{
\caption{\texttt{constructImage}$~\left(\textrm{dim}~d,~\textrm{size}~N,~\textrm{instance}~\varphi,~\textrm{type\_trait}~t \right)$, as utilized by Algorithm \ref{algo:highlevel} in line-5.\label{algo:constructImage}}
%\Header{\texttt{constructImage}$~\left(\textrm{dim}~d,~\textrm{size}~N,~\textrm{instance}~\varphi,~\textrm{type\_trait}~t \right)$}
%\hline
\Input{dimensionality $d$,\\sample-size $N$,\\function instance $\varphi$,\\type trait $t$,\\
}
\Output{a landscape image $\Gamma$ of type $t$ per function $\varphi$}
$\left\{ \vec{x}^{(j)}\right\}_{j=1}^{N} \leftarrow \textrm{Sample~} N \textrm{ feasible vectors in } \mathbb{R}^{d} \textrm{ uniformly at random}$\;
$M\leftarrow \texttt{extractFromTrait}\left( t \right)$\;
$\left\{ \vec{x}^{(j)}\right\}_{j=N+1}^{M} \leftarrow \textrm{Add~} (M-N) \textrm{ zero and/or basis vectors in } \mathbb{R}^{d} \textrm{ depending on } t$\;
$\left\{ f_j \right\}_{j=1}^{M} \leftarrow \texttt{evaluate}\left(\varphi, \left\{ \vec{x}^{(j)}\right\}_{j=1}^{M}\right)$\;
$\left\{ \vec{\tau}^{(j)}\right\}_{j=1}^{M} \leftarrow \texttt{concatenate}\left(f_j, \left\{ \vec{x}^{(j)}\right\}_{j=1}^{M}\right)$\;
$\Gamma \leftarrow \texttt{padAndForm}\left(t,~\left\{ \vec{\tau}^{(j)}\right\}_{j=1}^{M}\right)$; // this auxilary function is overloaded per each $t$\;
\Return{ $\Gamma$ }\;
%}
\end{algorithm*}
\DecMargin{2em}
% \begin{figure*}
% \centering
% \includegraphics[width=0.9\columnwidth]{precision_vs_d.eps}
% \includegraphics[width=0.92\columnwidth]{num_zeros.eps}
%  \caption{Detection using Type-1: test-set precision rates as a function of the problem dimensionality $d$ [LEFT], and as a function of $N$ per $d=22$, to reflect the effect of increasing sampling versus padding with zero vectors [RIGHT].\label{fig:accuracy}}
% \end{figure*}

\section{Experimental Setup and Results}\label{sec:results}
In this study, we target the automated detection of the noiseless test-functions of two benchmarking suites: BBOB \cite{hansen2016coco} (24 continuous functions) and IOHexperimenter \cite{DoerrYHWSB20} (23 discrete functions).

Importantly, we consider three levels of learning:
\begin{enumerate}[(L1)]
    \item Single-instance learning: each function is instantiated as a fixed random instance, that is, the learning algorithm is trained and tested over a single instance per function (i.e., taking line-4 in Algorithm \ref{algo:highlevel} outside the loop and switching it with line-3).
    \item Multi-instance learning: each function is instantiated as a fixed set of random instances, i.e., the algorithm is trained and tested over multiple fixed instances per function.
    \item Multi-unseen-instance learning: the training is conducted as in multi-instance learning, but the testing phase presents unseen instances to the algorithm.
\end{enumerate}
% we mainly focus on learning the benchmarking suite when each function is instantiated as a fixed random instance. In other words, the majority of our experiments considered a single instance per function (i.e., taking line-4 in Algorithm \ref{algo:highlevel} outside the loop and switching it with line-3). We will also discuss preliminary results of the generalized multi-instance detection in Section \ref{sec:discussions}.\\
%\paragraph{Data generation.} 
\textbf{Implementation: }\texttt{Python3} \cite{van1995python} with the PyTorch library \cite{NEURIPS2019_9015} were used for the CNN implementation (relying also on its tutorial's implementation \cite{PyTorchLecun98gradientbasedlearning}).
The data generator of the landscape images was written in \texttt{Python3}, partially relying on the public COCO \texttt{C}-code.

\subsection{Base Experiment: Single-Instance Detection using Type-1}
\begin{figure}
\centering
\includegraphics[width=0.9\columnwidth]{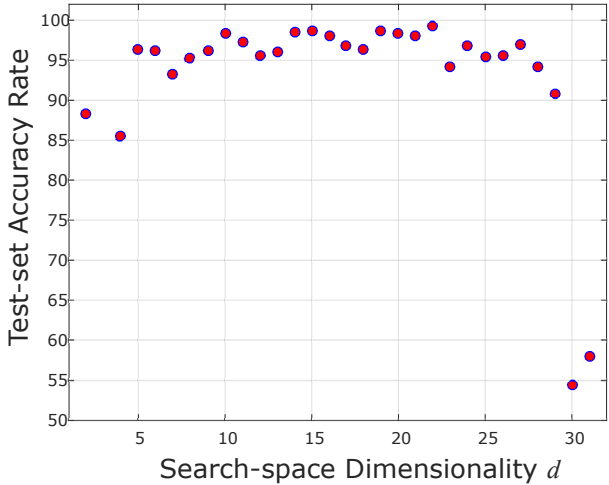}
\caption{Detection using Type-1: test-set accuracy rates as a function of the problem dimensionality $d$ \label{fig:precision_vs_d}}
\end{figure}

\begin{figure}
\centering
\includegraphics[width=0.92\columnwidth]{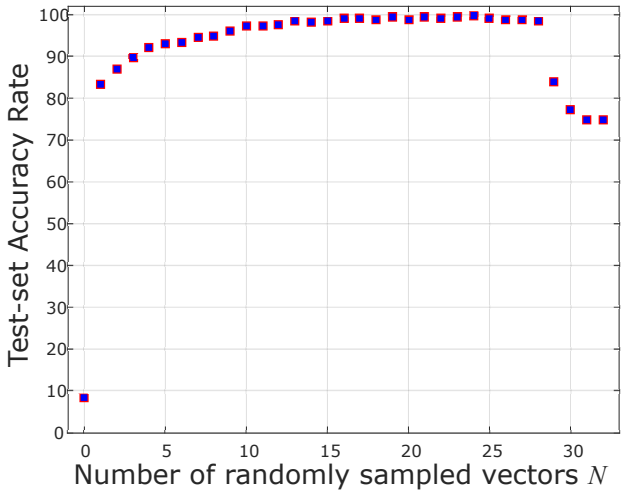}
 \caption{Detection using Type-1: test-set accuracy rates as a function of $N$ per $d=22$, to reflect the effect of increasing sampling versus padding with zero vectors.\label{fig:num_zeros}}
\end{figure}

We firstly addressed the L1 learning problem and trained a LeNet5 model for BBOB's 24 classes per a spectrum of search-space dimensions $d \in \left\{2,4,\ldots ,31 \right\}$. We start with the simplest landscape image structure, namely Type-1, in order to accomplish learning over all dimensions $d$ and explore the effectiveness of $N$. 
We trained this model over 3000 epochs with a learning rate of $10^{-6}$ and a batch size of 64, with one instance per function; the training and testing datasets were of sizes $24024$ and 11952, respectively.
we observed that the best accuracy rate was achieved for $d=22$ at epoch $2987$ (with accuracy rate $99.322\%$). Figure \ref{fig:precision_vs_d} depicts the best attained accuracy rates per each search-space dimension $d$. 
Furthermore, we explored the impact of $N$ for the case of $d=22$, and tested the model's performance when the number of sample vectors is increased, i.e. $N \in \left\{1,3,\ldots ,32 \right\}$. 
Following the aforementioned training setup, the best accuracy rate is observed for $N=24$ and at epoch $2859$ (accuracy rate $99.573\%$). Figure \ref{fig:num_zeros} depicts the best attained accuracy rates per each $N$ when $d=22$. 
%See also Table \ref{best_scores_func_breakdown}.


\subsection{Single-Instance Detection under Various Configurations}
We furthermore experimented the effect of setting up different configurations for the various landscape images.
Table \ref{best_scores_3000} presents selected use-cases and provides the best performance metrics when applying LeNet5 to them, with a learning rate $\alpha=10^{-6}$, batch size 64 and over 3000 epochs (training and testing datasets of 24024 and 11952 images, respectively). 
Two-letter abbreviations are used therein to denote various configurations. 
As previously explained, only Type-5 is capable of treating dimensions higher than 31. 

In light of the obtained results so far, we questioned whether even simpler neural models would be able to solve this classification problem. Especially, we wanted to question the necessity of the \textit{convolution} component to this end.
We thus addressed this function recognition problem using a single- and multi-layered Perceptron models \cite{Bishop}, replacing the LeNet5 CNN model. 
The multi-layered Perceptron consisted 3 layers. 
The parameters were set as above for type 1 landscape images (d=22 and N=24). 
Evidently, the models obtained accuracy rates of $95\%$ and $96\%$, respectively, exhibiting similar performance as the CNN model. 
This observation further emphasizes the efficiency of transferring the detection problem to the computer vision field, and suggests that the image recognition problem is `easy' to some extent.

Furthermore, function recognition subject to additive Gaussian noise, when applied directly to the landscape images, was also tested for the LeNet5 CNN. 
The amplitude of this additive noise was calculated as a Gaussian distribution from zero until half-image-maximum. 
The parameters were set as above, except for setting $\alpha=10^{-5}$ and 1000 epochs for training. 
The obtained accuracy rate for one instance per function was $88\%$. 
This accuracy rate is close to the noiseless results, indicating the robustness to noise of the proposed method.
Notably, applying additive noise to landscape images, as reported herein, is not comparable to treatment of noisy optimization problems, where distinction is made between noisy input (i.e., decision variables' vector) to noisy output (i.e., the objective function value) -- see, e.g., \cite{Shir-MOQC}.

\begin{table}[h]
\caption{Best performance metrics over the noiseless COCO suite when employing LeNet5 with a learning rate $\alpha=10^{-6}$, batch size 64 and over 3000 epochs. Training, validation and testing datasets were of sizes $24024$, $12024$, and $11952$ images, respectively.
\label{best_scores_3000}}
\centering
\begin{small}
\begin{tabular}{lccccc}
\toprule
\thead{Image\\Type} &
  \thead{Config\\ ID} &
  \thead{Dim. \\ $d$} &
  \thead{Train set\\ precis.} &
  \thead{Validation\\  precis.} &
  \thead{Test set\\ precis.} \\
\midrule
Type-3 & uo & 22 &   94.481 & 94.286 & 94.227 \\
Type-5 & an & 22 &  92.233 & 92.041 & 91.809 \\
Type-4 & uc & 22 & 89.244 & 88.706 & 89.232 \\
Type-2 & oc & 4 & 79.579 & 79.291 & 79.359\\
Type-5 & an & 40 & 100 & 100 & 100\\
Type-3 & uo & 4 & 83.304 & 82.585 & 82.764\\
Type-4 & uc & 4 & 78.813 & 79.034 & 78.455\\
Type-1 & oo & 4 & 80.191 & 79.316 & 79.911\\
Type-2 & oc & 22 & 91.017 & 90.569 & 90.504\\
Type-5 & an & 4 & 80.261 & 80.281 & 80.313\\
Type-1 & oo & 22 & 95.45 & 95.434 & 95.214 \\ 
\bottomrule
\end{tabular}
\end{small}
\end{table}

\subsection{Comparative Single-Instance Detection by all Image Types}
\begin{figure}
\centering
\includegraphics[width=0.8\columnwidth]{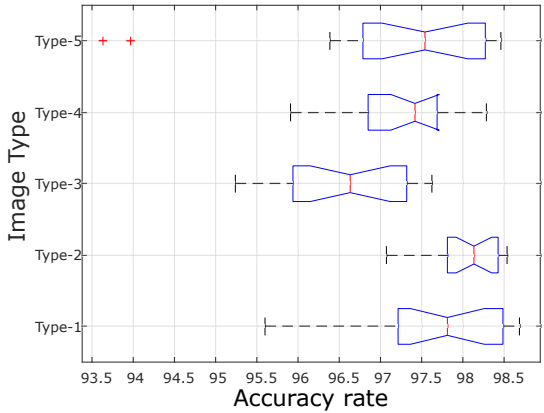}
 \caption{Statistical boxplots accounting for 20 runs of the detection problem at dimension $d=22$ using the 5 images types with $N=24$. 
 The compared performance measure is the test-set accuracy rate of the model with the minimal loss per run. LeNet5 is employed with a learning rate $\alpha=10^{-5}$, batch size 64 and is run over 800 epochs, training and testing datasets of $24024$ and $11952$ images, respectively. \label{fig:boxplots_d22}}
\end{figure}

We conducted comparisons among the various image types, aiming to assess their relative performance in different setups over the L1 problem. 
While the comparisons are not conclusive in terms of a statistically-significant ranking, some interesting, problem-dependent trends are evident -- as will be further discussed in Section \ref{sec:discussions}.
Figure \ref{fig:boxplots_d22} provides the statistical boxplots for comparing the 5 image types on the COCO classification problem at dimension $d=22$ (reflecting 20 runs).  

\subsection{Addressing the Multi-Instance Problem (L2)}
%\td{TODO: update/verify the following text}

\begin{figure}
\centering
\includegraphics[width=1\columnwidth]{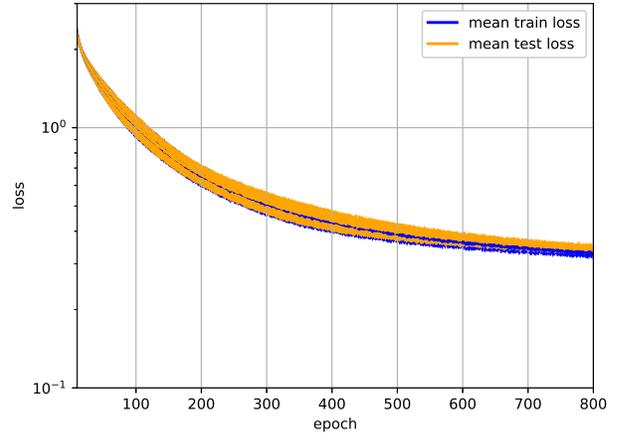}
 \caption{Averaged \textit{loss} over 20 runs for the multi-instance detection problem at dimension $d=22$ using Type-1 images with $N=24$.
 LeNet5 is employed with a learning rate $\alpha=10^{-5}$, batch size 64 and is run over 800 epochs. The training and testing datasets include $30120$ and $14760$ images, respectively. The standard deviation among runs is depicted as the shaded areas surrounding each curve.\label{fig:loss_instances}}
\end{figure}

We also conducted experiments of the generalized multi-instance detection problem (i.e., realizing Algorithm \ref{algo:highlevel} as is).
Preliminary results of such experiments indicate that the generalized problem is learnable as well, but the existing configurations of LeNet5 obtain lower performance measures. 
For example, the learning process on 20 datasets with 5 possible instances per function is presented in Figure \ref{fig:loss_instances}, depicting the \textit{averaged loss} for this preliminary use-case. The averaged testing accuracy for the best nets trained on each of these 20 datasets was $84.71\%$ with a standard deviation of $60.19\%$. 
Finally, Table \ref{best_scores_3000} provides the mean accuracy, precision, recall and F1 measures for the explicit functions' break-down (considering 5 instances for testing the best net). 

%The global mean measures for these run were as follows: $accuracy=0.8534\pm0.0019$, $precision=0.8534\pm0.0019$, $recall=0.8534\pm0.0019$ and $F1=0.8534\pm0.0019$. 

\begin{table*}[htbp]
\caption{Best performance metrics over the noiseless COCO suite using LeNet5: functions' break-down. The functions are partitioned into their 5 organic classes (i-v, as specified in Section \ref{sec:introduction}); see also  \cite{Hansen2014Real}.
The measures per each function comprise the mean values (first line), followed by the standard deviations. All the reported results are for the following setup: 20 testing datasets each of 30120 images, net learning dataset of $14760$ images, search-space dimensionality $d=22$, $N=24$, Type-1, with 5 different instances per function; learning rate $\alpha=10^{-5}$, $800$ epochs, batch size $64$.}
\label{best_scores_func_breakdown}
%\scalebox{0.77}{
\centering
\begin{small}
\begin{tabular}{lcccccc}
\toprule
%{l *{8}{p{1cm}}
\multicolumn{1}{l}{Function} &
  %\thead{Dim. \\ $d$} &
  \thead{Test set\\ accuracy} &
  \thead{Test set\\ precision} &
  \thead{Test set\\  recall} &
  \thead{Test set\\ F1-score} \\ % &
  %Epoch &
   % Image Type  \\
\midrule
1. Sphere (Mean) & 0.9827 & 0.9862 & 0.9827 & \textbf{0.9844} \\ %& 6 & 1  \\ 
\hspace{1.4cm} (STD) & 0.0032 & 0.0033 & 0.0032 & 0.002 \\ [0.1cm] %\hline
2. Ellipsoidal A (Mean) & 0.6912 & 0.6375 & 0.6912 & \textbf{0.6632} \\ 
\hspace{1.4cm} (STD) & 0.016 & 0.0128 & 0.016 & 0.0135 \\ [0.1cm] %\hline
3. Rastrigin (Mean) & 0.6912 & 0.6247 & 0.6912 & \textbf{0.6562} \\ 
\hspace{1.4cm} (STD) & 0.013 & 0.0111 & 0.013 & 0.011 \\ [0.1cm] %\hline
4. Rastrigin-B{\"u}che (Mean) & 0.9404 & 0.9603 & 0.9404 & \textbf{0.9502} \\ 
\hspace{1.4cm} (STD) & 0.0068 & 0.0042 & 0.0068 & 0.0041 \\ [0.1cm] %\hline
5. Linear Slope (Mean) & 0.9998 & 0.9993 & 0.9998 & \textbf{0.9996} \\ 
\hspace{1.4cm} (STD) & 0.0004 & 0.0005 & 0.0004 & 0.0003 \\ %[0.1cm] %\hline
\midrule
6. Attractive Sector (Mean) & 0.8666 & 0.873 & 0.8666 & \textbf{0.8698} \\ 
\hspace{1.4cm} (STD) & 0.0063 & 0.0083 & 0.0063 & 0.0049 \\ [0.1cm] %\hline
7. Step Ellipsoidal (Mean) & 0.6823 & 0.5691 & 0.6823 & \textbf{0.6205} \\ 
\hspace{1.4cm} (STD) & 0.0148 & 0.0088 & 0.0148 & 0.0097 \\ [0.1cm] %\hline
8. Rosenbrock (Mean) & 0.824 & 0.6241 & 0.824 & \textbf{0.7102} \\ 
\hspace{1.4cm} (STD) & 0.0113 & 0.0075 & 0.0113 & 0.0078 \\ [0.1cm] %\hline
9. Rotated Rosenbrock (Mean) & 0.9974 & 0.996 & 0.9974 & \textbf{0.9967} \\ 
\hspace{1.4cm} (STD) & 0.001 & 0.0018 & 0.001 & 0.001 \\ [0.1cm] %\hline
\midrule
10. Ellipsoidal B (Mean) & 0.6597 & 0.7173 & 0.6597 & \textbf{0.6872} \\ 
\hspace{1.4cm} (STD) & 0.0136 & 0.0143 & 0.0136 & 0.0127 \\ [0.1cm] %\hline
11. Discus (Mean) & 0.7919 & 0.8384 & 0.7919 & \textbf{0.8145} \\ 
\hspace{1.4cm} (STD) & 0.0087 & 0.0101 & 0.0087 & 0.0079 \\ [0.1cm] %\hline
12. Bent Cigar (Mean) & 0.8765 & 0.826 & 0.8765 & \textbf{0.8505} \\ 
\hspace{1.4cm} (STD) & 0.0093 & 0.0061 & 0.0093 & 0.0061 \\ [0.1cm] %\hline
13. Sharp Ridge (Mean) & 0.9996 & 0.9988 & 0.9996 & \textbf{0.9992} \\ 
\hspace{1.4cm} (STD) & 0.0005 & 0.0009 & 0.0005 & 0.0005 \\ [0.1cm] %\hline
14. Different Powers (Mean) & 0.8443 & 0.8111 & 0.8443 & \textbf{0.8273} \\ 
\hspace{1.4cm} (STD) & 0.0092 & 0.0084 & 0.0092 & 0.0059 \\ [0.1cm] %\hline
\midrule
15. Rastrigin (Mean) & 0.392 & 0.568 & 0.392 & \textbf{0.4637} \\ 
\hspace{1.4cm} (STD) & 0.0089 & 0.0107 & 0.0089 & 0.0068 \\ [0.1cm] %\hline
16. Weierstrass (Mean) & 0.9982 & 0.9994 & 0.9982 & \textbf{0.9988} \\ 
\hspace{1.4cm} (STD) & 0.0011 & 0.0005 & 0.0011 & 0.0005 \\ [0.1cm] %\hline
17. Schaffers F7 (Mean) & 0.9709 & 0.9705 & 0.9709 & \textbf{0.9707} \\ 
\hspace{1.4cm} (STD) & 0.0053 & 0.0055 & 0.0053 & 0.0043 \\ [0.1cm] %\hline
18. Schaffers F7 moderately ill-conditioned (Mean) & 0.7869 & 0.8251 & 0.7869 & \textbf{0.8054} \\ 
\hspace{1.4cm} (STD) & 0.0133 & 0.0098 & 0.0133 & 0.0075 \\ [0.1cm] %\hline
19. Composite Griewank-Rosenbrock F8F2 (Mean) & 0.9878 & 0.9788 & 0.9878 & \textbf{0.9833} \\ 
\hspace{1.4cm} (STD) & 0.002 & 0.0043 & 0.002 & 0.0022 \\ [0.1cm] %\hline
\midrule
20. Schwefel  (Mean) & 0.5014 & 0.7177 & 0.5014 & \textbf{0.5903} \\ 
\hspace{1.4cm} (STD) & 0.0139 & 0.0137 & 0.0139 & 0.0124 \\ [0.1cm] %\hline
21. Gallagher’s Gaussian 101-me Peaks (Mean) & 0.9997 & 0.9974 & 0.9997 & \textbf{0.9985} \\ 
\hspace{1.4cm} (STD) & 0.0007 & 0.0012 & 0.0007 & 0.0007 \\ [0.1cm] %\hline
22. Gallagher’s Gaussian 21-hi Peaks (Mean) & 0.9976 & 0.9999 & 0.9976 & \textbf{0.9987} \\ 
\hspace{1.4cm} (STD) & 0.0012 & 0.0003 & 0.0012 & 0.0007 \\ [0.1cm] %\hline
23. Katsuura (Mean) & 0.9994 & 0.9993 & 0.9994 & \textbf{0.9994} \\ 
\hspace{1.4cm} (STD) & 0.0005 & 0.0008 & 0.0005 & 0.0005 \\ [0.1cm] %\hline
24. Lunacek bi-Rastrigin (Mean) & 0.9997 & 0.9992 & 0.9997 & \textbf{0.9995} \\ 
\hspace{1.4cm} (STD) & 0.0005 & 0.0009 & 0.0005 & 0.0005 \\ [0.1cm] %\hline
\bottomrule
\end{tabular}
\end{small}
%}
\end{table*}

\subsection{Addressing the Multi-Unseen-Instance Problem (L3)}
\begin{figure}
\centering
\includegraphics[width=0.8\columnwidth]{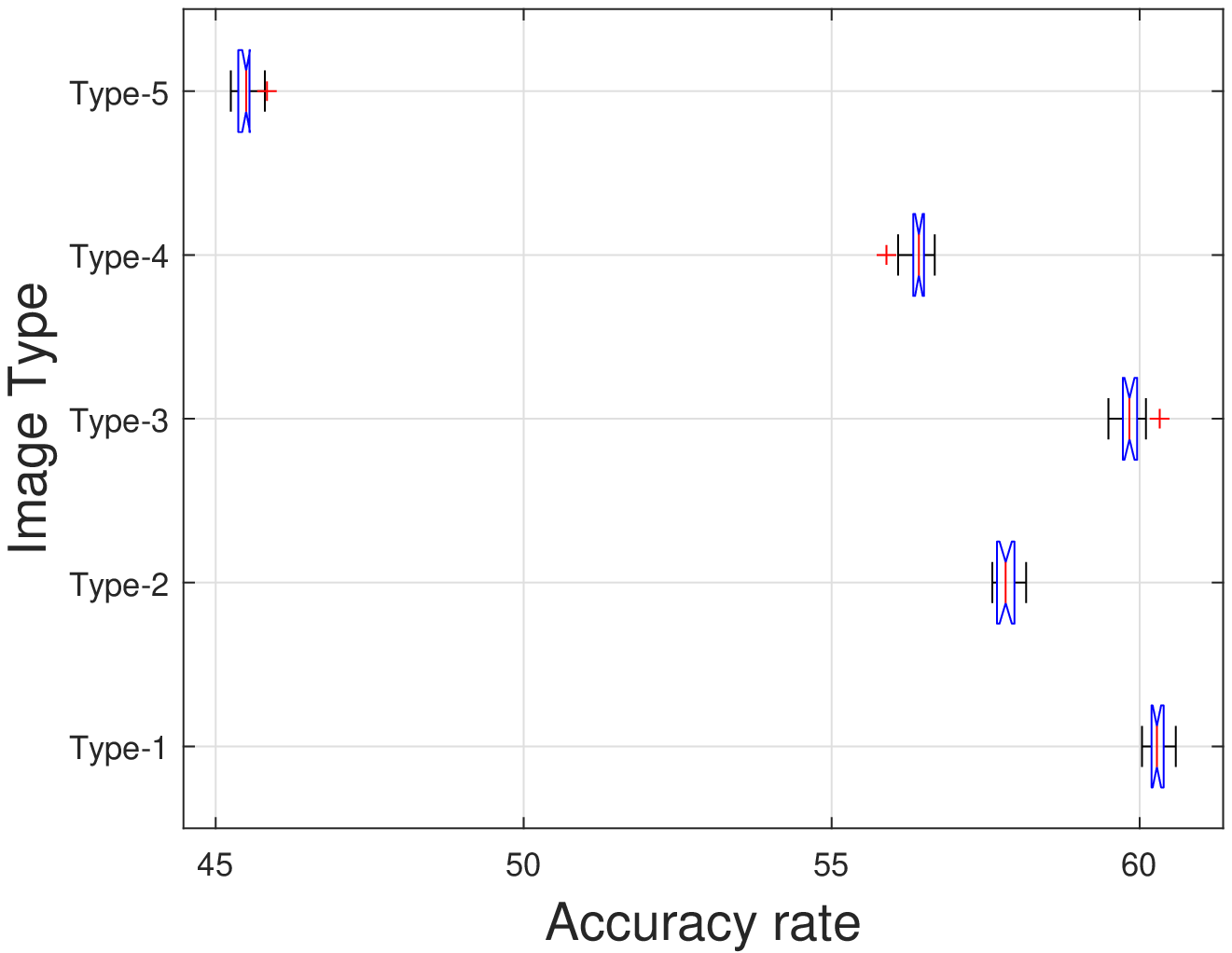}
 \caption{Statistical boxplots accounting for 20 runs of the Multi-Unseen-Instance Problem at dimension $d=22$ using the 5 image types with $N=24$; noise-free data.
  \label{fig:boxplots_d22_1}}
\end{figure}

\begin{figure}
\centering
\includegraphics[width=0.8\columnwidth]{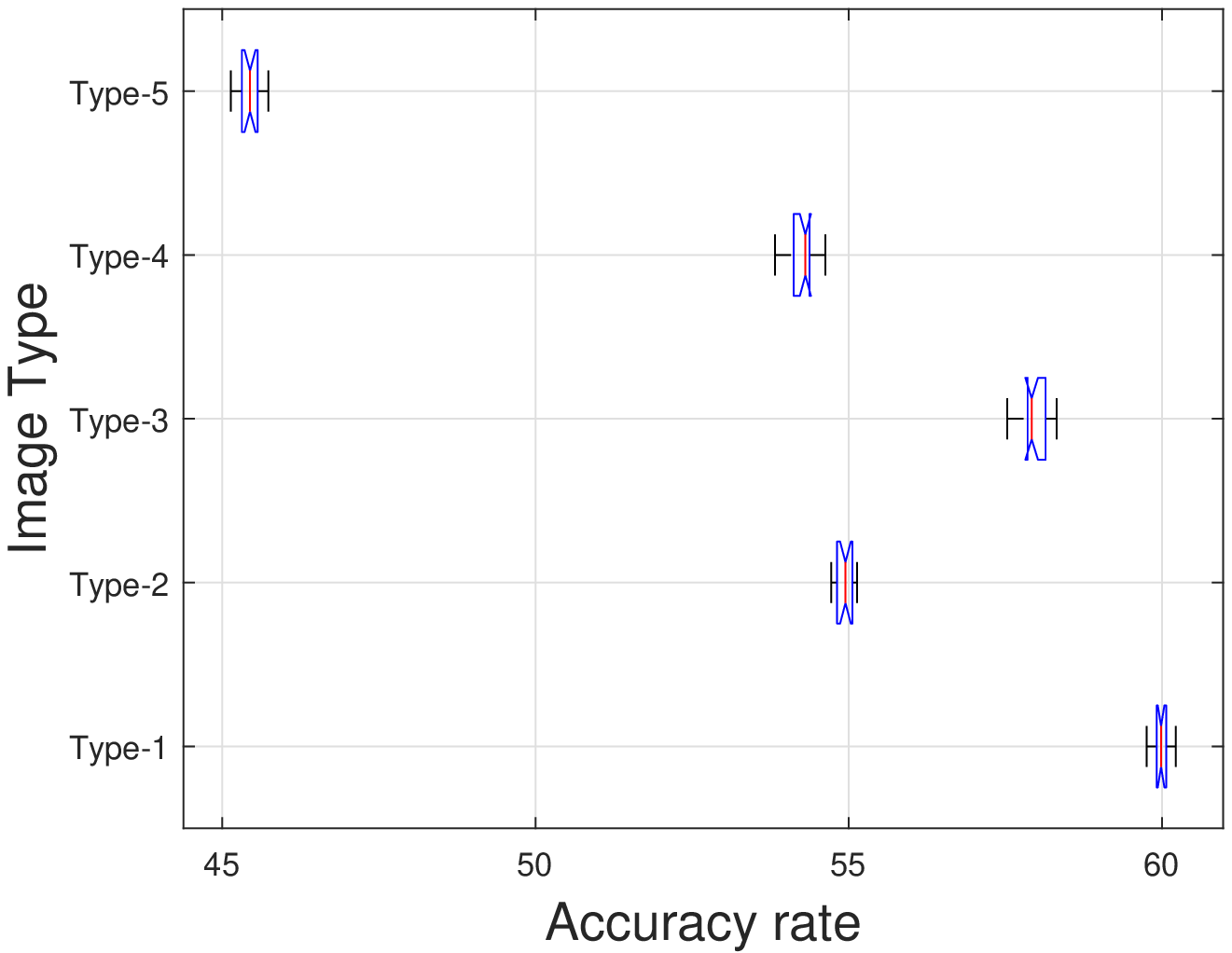}
 \caption{Statistical boxplots accounting for 20 runs of the Multi-Unseen-Instance Problem at dimension $d=22$ using the 5 image types with $N=24$; additive Gaussian noise was applied to the images. 
  \label{fig:boxplots_d22_2}}
\end{figure}

In case of Multi-Unseen-Instance Problem we tested detection of the best nets trained and tested over BBOB - 24 functions with 5 instances datasets of 30120 and 14760 images, respectively. 
The learning process was accomplished for each landscape image type by the LeNet5 model, with a learning rate $\alpha=10^{-6}$, batch size 64 and over 3000 epochs. Pre-trained functions were applied at 20 datasets of 30120 images generated by 5 different instances of the same functions. In addition, uniform noise in the range of $[-2.5, 2.5]$ was added to the data (to the whole Landscape image). 
The data range was $[5, 5]$. The detection accuracy rate for the best performing landscape image type 1 was on average $60.289\%$ with a standard deviation of $0.154\%$ and $59.988\%$ with a standard deviation of $0.118\%$, for noiseless and noisy cases, respectively. 
Boxplots for each landscape image type are presented in Figures \ref{fig:boxplots_d22_1} and \ref{fig:boxplots_d22_2} for the noiseless and noisy settings, respectively.

\subsection{Single-Instance Detection for the IOHexperimenter Data}
The generated data for IOHexperimenter's single-instance 21 functions was organized as type 1 landscape images. The LeNet5 model, using a learning rate of $\alpha=10^{-6}$, batch size of 64, and utilizing over 3000 epochs (training and testing datasets of 21021 and 10458 images, respectively), was applied to the data. Following 3000 epochs of training, the model obtained training and testing average accuracy rates of $25.26\%$ and $24.029\%$, respectively. 
The model achieved at its peak (after 2590 epochs of training) a testing average accuracy rate of $23.704\%$. The results per-functions, for the model at its peak, are presented in Table \ref{best_scores_func_IOHexperimenter}.

\begin{table} [h]
\caption{Function recognition results for the \\ best net, over IOHexperimenter \\ suite using LeNet5.}
\label{best_scores_func_IOHexperimenter}
%\scalebox{0.77}{
\centering
\begin{small}
\begin{tabular}{lcccccc}
\toprule
%{l *{8}{p{1cm}}
%\column{1}{l}{Function} &
  %\thead{Dim. \\ $d$} &
  \thead{Test set\\ accuracy} \\ % &
  %Epoch &
   % Image Type  \\
\midrule
1. OneMax (Mean) & $9\%$ \\
2. LeadingOnes A & $28\%$ \\ 
3. Linear & $0\%$ \\ 
4. OneMaxDummy1 & $12\%$ \\ 
5. OneMaxDummy2 & $26\%$ \\ 
6. OneMaxNeutrality & $7\%$ \\ 
7. OneMaxEpistasis & $4\%$ \\ 
8. OneMaxRuggedness1 & $29\%$ \\ 
9. OneMaxRuggedness2 & $1\%$ \\ 
10. OneMaxRuggedness3 & $10\%$ \\ 
11. LeadingOnesDummy1 & $7\%$ \\ 
12. LeadingOnesDummy2 & $4\%$ \\ 
13. LeadingOnesNeutrality & $3\%$ \\ 
14. LeadingOnesEpistasis  & $2\%$ \\ 
15. LeadingOnesRuggedness1 & $4\%$ \\ 
16. LeadingOnesRuggedness2 & $26\%$ \\ 
17. LeadingOnesRuggedness3 & $23\%$ \\ 
18. LABS & $100\%$ \\ 
19. MIS & $0\%$ \\ 
20. IsingRing  & $100\%$ \\ 
21. IsingTriangular & $100\%$ \\ 
\bottomrule
\end{tabular}
\end{small}
%}
\end{table}
\section{Discussion}\label{sec:discussions}	
Evidently, the noiseless BBOB and IOHexperimenter suites are learnable by a basic CNN \cite{Lecun98gradient-basedlearning}. 
The neural learning model was capable of attaining efficient function recognition, using a carefully-designed database and a moderate number of learning iterations.
Figures \ref{fig:precision_vs_d}-\ref{fig:num_zeros} exhibit fine performance of the learning model for a wide range of $d$ and $N$. 
The classification scores deteriorate only towards the end of the range. 
This behavior may be explained by the CNN's pooling effects, which diminish the influence of the least repeated data. For example, for $d=30$, there are only 2 copies of sample vectors within our setup of $M=32$, which can be lost due to pooling.

When comparing the image types, we are unable to make statistically-significant statements across all setups and configurations. 
However, it is evident in Figure \ref{fig:boxplots_d22} that Type-2 performs best on average per the reported use-case therein. 
This can be explained by the constructed replication of $\varphi_{k,i}\left(\vec{x}^{(j)}\right)$ and $\vec{x}^{(j)}$. 
This coupled replication seemingly preserves more structural information about the functions' response surfaces, being beneficial for the BBOB suite.
%This result indicates that it is preferred to replicate only $\varphi_{k,i}\left(\vec{x}^{(j)}\right)$ to provide it with more influence during CNN operations.

The preliminary multi-instance results exhibit accuracy rates' deterioration to $84.71\%$, $60.289\%$ and $59.988\%$, for multi-instance, multi-instance-unseen and multi-instance-unseen-noisy conditions, respectively. 
Yet, the accuracy rates are relatively high and robust even in noisy and unlearned conditions. 
These results can be improved using more efficient configurations, and especially by a broader and a more flexible CNN model.  

Table \ref{best_scores_3000} presents fine accuracy, precision, recall and F1 measures for most of the functions. However, there are certain functions for which the recognition was less effective -- in particular, the Rastrigin (classes i and iii), Step-Ellipoid (class ii), and the Schwefel function (class v). These observations may be explained by the highly multimodal global structure of the Rastrigin landscape, the multiplicity of plateaus within the Step-Ellipsoid landscape, and the weak global structure of the Schwefel landscape.
At the same time, further investigation is much needed to explain this interesting behavior.

%In our feature work we plan to research reasons and solutions for this behaviour of the functions,
The results for the IOHexperimenter suite (Table \ref{best_scores_func_IOHexperimenter}) show on average lower detection rate of about $24\%$, whereas certain functions within this suite enjoy better detection rates of up to $100\%$. 
Notably, the results indicate that learning was accomplished, being better than the random classification hypothesis. The low detection rates may be attributed to te complex nature of the discrete landscapes.
A future direction of research will aim to explain the differences among the detection rates in this suite. In addition, landscape images and a suitable learning model will be developed to address the IOHexperimenter suite. 

\subsection*{Conclusions and Future Work}
In this study we formulated the continuous BBO benchmarking-suite detection problem as an image recognition challenge. We reported on successfully addressing the multi-instance detection problem in high dimensions ($\lesssim 40$) per the BBOB and IOHexperimenter suites. %and also presented preliminary results on the generalized multi-instance detection problem. %Per the generalized problem, improving the results likely requires the modification of the CNN. % and/or further investigating the configurations. 
This novel perspective of representing high-dimensional continuous search landscapes as plain images, and the evident success in learning them, pave altogether the way toward automated feature extraction in BBO problems.
Notably, once a robust model capable of learning the generalized problem is attained, even in the cost of a large number of objective function calls/queries during the training phase, the model will encompass the search-landscapes features.
Consequently, during deployment -- e.g., when addressing the \textit{algorithm selection problem} -- unseen BBO problem-instances will be then required to provide a single image per each local phase of the search, which translates into an order of $d$ objective function calls/queries.

One direction of future research is the investigation of the obtained neural networks in order to explore the underlying features by unveiling the convolution information. Should this effort become fruitful, the machine-based features are to be interpreted and compared with the human experts' features, e.g., as used within the ELA approaches.

From the engineering point of view, the proposed methodology can be used to recognize loss functions of different optimization and learning problems. 
That is, the established model will be able to recognize the closest image, within the set of known loss functions' images, that fits most the target function. Such a set of known images will play the equivalent role of ImageNet for optimization problems' detection.
Subsequently, it will be possible to fit algorithms to a given problem at hand, which in turn will enable smoother problem-solving. %with its' matching loss function. It in turn will provide more efficient optimization and learning. 
In addition, our method can be used to recognize mathematical functions describing natural and engineering processes and structures. 
For example, an airplane wing's condition can be described as a function of its fluctuations as well as another function of its load. 
Our method will be able to recognize the fluctuation function using input-load and output-fluctuation. Once the function is recognized, one can determine the structural condition of the wing. %Wing here viewed as a system described by the function. 
Another idea would be to describe structures with functions. For example, in vocal instruments, the structure is the mapping of the playing notes to the sound.

Additional future pathways of research comprise the experimentation of additional BBO landscapes and test-suites, consideration of noise (at the input as well as the output levels), sampling strategy, multi-instance setups, and addressing scalability questions (by resizing $M$ respectively), also in light of the renowned \textit{curse of dimensionality}. 
%Results of the study can be also applied to other fields of science and engineering. For example imagine recognition of vibration function of the airplane engine and possible diagnostics tools based on it.

%\section*{Acknowledgments}
% The authors are indebted to Ido Dan, for his preliminary contributions during the ignition of this research.

\bibliographystyle{ieeetr}
\bibliography{DLreferences}
%\bibliography{references1}
\end{document}